\documentclass[letterpaper]{article} 
\usepackage[draft]{aaai25}
\usepackage{times}  
\usepackage{helvet}  
\usepackage{courier}  
\usepackage[hyphens]{url}  
\usepackage{graphicx} 
\usepackage{natbib}  
\usepackage{caption} 
\usepackage{algorithm}
\usepackage{algorithmic}

\usepackage{newfloat}
\usepackage{listings}
\DeclareCaptionStyle{ruled}{labelfont=normalfont,labelsep=colon,strut=off} 
\floatstyle{ruled}
\newfloat{listing}{tb}{lst}{}
\floatname{listing}{Listing}
\usepackage{amsmath,amsfonts}
\usepackage{array}
\usepackage{textcomp}
\usepackage{url}
\usepackage{verbatim}
\usepackage{multirow}
\usepackage{color}
\usepackage{marvosym}
\usepackage{booktabs}

\definecolor{url_color}{RGB}{42, 83, 163}

\DeclareMathOperator{\Exp}{Exp}

\DeclareMathOperator{\diag}{diag}
\newcommand{\norm}[1]{\left\lVert#1\right\rVert}

\newcommand{\ywc}[1]{\textcolor{black}{#1}}
\newcommand{\gxy}[1]{\textcolor{black}{#1}}

\title{D$^3$FlowSLAM: Self-Supervised Dynamic SLAM with \\Flow Motion Decomposition and DINO Guidance}
\author{
    Xingyuan Yu\textsuperscript{\rm 1,*},
    Weicai Ye\textsuperscript{\rm 1,}\thanks{Equal Contribution.},
    Xiyue Guo\textsuperscript{\rm 1},
    Yuhang Ming\textsuperscript{\rm 2},\\
    Jinyu Li\textsuperscript{\rm 3},
    Hujun Bao\textsuperscript{\rm 1},
    Zhaopeng Cui\textsuperscript{\rm 1},
    Guofeng Zhang\textsuperscript{\rm 1}\thanks{Corresponding Author.}
}
\affiliations{

    \textsuperscript{\rm 1}State Key Lab of CAD \& CG, Zhejiang University
    \textsuperscript{\rm 2}Hangzhou Dianzi University
    \textsuperscript{\rm 3}Dreame Technology\\
}

\begin{document}

\maketitle

\begin{abstract}
    \gxy{In this paper, we introduce a self-supervised deep SLAM method that robustly operates in dynamic scenes while accurately identifying dynamic components. Our method leverages a dual-flow representation for static flow and dynamic flow, facilitating effective scene decomposition in dynamic environments. We propose a dynamic update module based on this representation and develop a dense SLAM system that excels in dynamic scenarios. In addition, we design a self-supervised training scheme using DINO as a prior, enabling label-free training. Our method achieves superior accuracy compared to other self-supervised methods. It also matches or even surpasses the performance of existing supervised methods in some cases.
    All code and data will be made publicly available upon acceptance.
    }
\end{abstract}

%

\section{Introduction}
\label{sec:intro}

\begin{figure*}[t]
    \centering
    \includegraphics[width=1\linewidth]{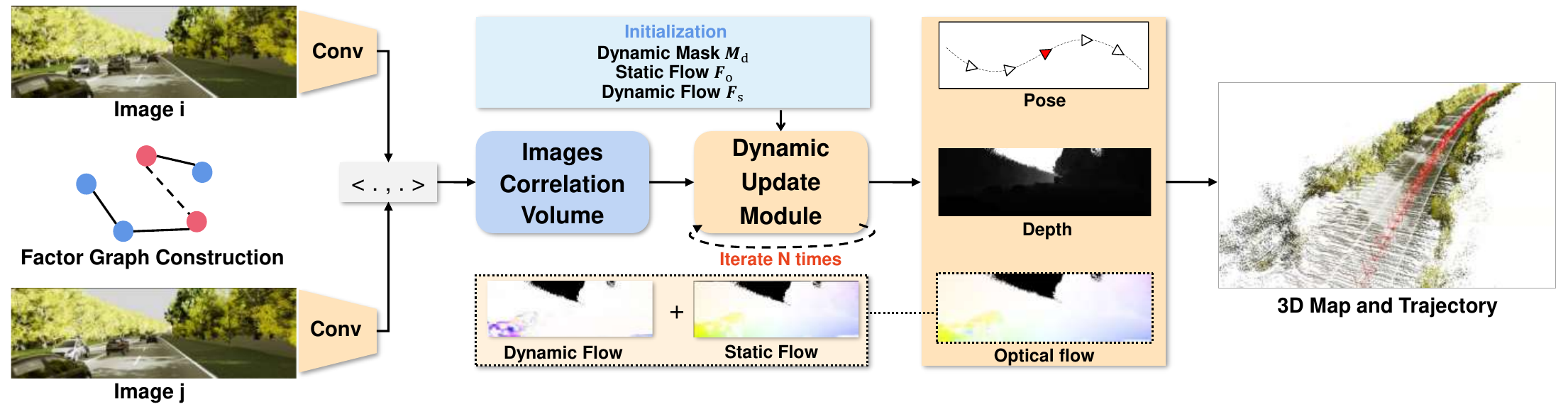} 
    \caption{\textbf{D$^3$FlowSLAM Overview.} D$^3$FlowSLAM takes an image sequence as input, extracts features to construct a correlation volume, and then combines this with the initial static flow, dynamic flow, and dynamic mask before feeding it into the dynamic update module. This module iteratively optimizes the residuals of pose, inverse depth, static flow, dynamic flow, and dynamic mask, ultimately providing estimates of the camera pose, 3D structure, and dynamic decomposition results.
    }
    \label{fig:pipeline}
\end{figure*}

Simultaneous localization and mapping (SLAM) is fundamental to the fields of computer vision and robotics, with applications ranging from augmented reality (AR) and virtual reality (VR) to autonomous driving. In AR, SLAM is commonly used for accurate localization of agents, enabling users to place virtual objects~\cite{Ye2021SuperPlane}, while dense reconstruction is crucial for better interaction with the surrounding environment. Due to the simplicity of monocular video acquisition, monocular dense SLAM~\cite{deepfactors, tateno2017cnn} has attracted significant attention, although it remains a more complex task compared to RGB-D SLAM~\cite{wang2021DymSLAM, Li2022DGS-SLAM, yu2018dsslam, dai2020rgb, henein2020dynamic}. Significant progress has been made in sparse traditional~\cite{Engel2014lsd, mur2015orb, forster2014svo, engel2017direct}, learning-based~\cite{Zhou2017SfMLearner, wang2018end, wang2019improving}, and hybrid~\cite{wang2017non, wang2017noniterative, brahmbhatt2018geometry} approaches. More recently, a learning-based monocular dense SLAM system, DROID-SLAM~\cite{teed2021droid}, has been proposed. It demonstrates better accuracy and robustness than other methods but requires \textbf{extensive labeled data}, such as camera poses and depths, during training, which limits its \textbf{adaptability for new scenes}.

Developing robust SLAM methods for real-world AR applications presents significant challenges, particularly in dynamic environments. For instance, we find that dynamic objects can create ambiguity during flow residual estimation in~\cite{teed2021droid}, which often results in suboptimal performance. To address these challenges, some approaches~\cite{bescos2018dynaslam, xiao2019dynamic} pre-filter dynamic objects using segmentation before executing monocular sparse SLAM. Other strategies~\cite{yang2019cubeslam, huang2020clustervo} integrate object detection with SLAM to mitigate these issues. However, these methods often encounter limitations in practical scenarios, as the system's generalization capability is restricted by the training data of the detector, leaving some dynamic objects unaddressed~\cite{wang2021DymSLAM}. If the detector generates inaccurate outputs, the system may fail entirely. Furthermore, many of these approaches are tightly coupled with sparse SLAM, with limited focus on dense SLAM. Directly ignoring dynamic information and concentrating solely on sparse systems can lead to the loss of valuable data.

Building on previous work, we argue that a dense dual-flow representation can effectively address these challenges. This approach offers several benefits: 1) it optimizes poses and depths using static flows in a standard manner, 2) it learns dynamic flows by maintaining consistent luminosity through warping the current frame to adjacent frames, and 3) it fully utilizes information from all pixels during estimation. Furthermore, this flow-based representation facilitates the development of a \textbf{self-supervised training scheme} for our model. With these concepts in mind, we introduce D$^3$FlowSLAM, a dense method that incorporates a dynamic update module based on dual-flow representation, as shown in Fig.~\ref{fig:pipeline}. We leverage the self-supervised model DINO \cite{caron2021emerging} to obtain a prior foreground mask by clustering its features~\cite{amir2021deep}. Then, in the update module, we employ a dynamic mask branch guided by two priors: the foreground mask generated from DINO features and a mask derived from the dual-flow difference. This update process is detailed in Fig.~\ref{fig:update module}.

Our method effectively harnesses information from different regions of the pixels, resulting in more accurate estimations in dynamic scenes. Lastly, we design a \textbf{self-supervised training scheme} that enables the exploitation of an unlimited amount of \textbf{unlabeled data} and \textbf{finetuning on new images} without labels. Compared to the latest self-supervised method \cite{sun2024dynamo} for dynamic scenes, our approach yields superior results and significantly outperforms it in SLAM tasks.

 In summary, our contributions are three-fold:
\begin{itemize}
    \item We introduce a dynamic update module based on dual-flow representation, enabling it to handle dynamic and static parts of the scene in an end-to-end fashion.
    \item We use our model to develop a dense dynamic SLAM system,
    which can perform robustly in dynamic scenes.
    \item 
    We propose a self-supervised training scheme that yields the best results among self-supervised methods, with performance comparable to supervised methods.
\end{itemize}



\section{Related Work}

\subsection{Dynamic SLAM}
Humans live in a dynamic environment, intelligent systems should also have the ability to deal with dynamic environments, recognizing the dynamic contents from the static environments. 
Traditional approaches primarily mitigate the interference of dynamic objects by introducing a prior~\cite{tan2013robust} or RANSAC methods~\cite{huang2020dual}. Some recent approaches try to use segmentation methods like \cite{He_2017_ICCV} to filter out potential dynamic objects like vehicles and pedestrians~\cite{vincent2020dynamic, bescos2018dynaslam, xiao2019dynamic} and then run the sparse SLAM system like \cite{mur2015orb},
or unify object detection and SLAM into a multi-task system \cite{huang2020clustervo, nair2020multi, ye2023pvo}, or add the object constraint to the SLAM system~\cite{yang2019cubeslam, strecke2019fusion, brasch2018semantic} to eliminate the interference of dynamic objects. 
Overall, the aforementioned methods mainly address the problem of localization and mapping in dynamic scenes by adding auxiliary modules to traditional sparse SLAM systems. In contrast, our method takes an end-to-end approach, integrating the identification of dynamic regions and pose optimization within a single dense framework for joint optimization. Similar to \cite{zhan2021dfvo,teed2021droid}, our learning-based approach can identify dynamic fields at the pixel level, allowing for more effective use of image information and better simulating how humans perceive the world.

\subsection{Self-Supervised Learning in Vision}
Self-supervised learning has recently advanced in computer vision, enabling models to autonomously learn high-quality features from large image datasets~\cite{he2020momentum,he2022masked,oquab2023dinov2,caron2021emerging}. With minimal fine-tuning, these models can achieve strong performance in downstream tasks. Additionally, leveraging their features allows for human-interpretable results through simple clustering or operations without parameter updates~\cite{amir2021deep}.
In specific tasks like optical flow estimation, the predictions of self-supervised methods~\cite{meister2018unflow,luo2021upflow} have progressively approached the accuracy of supervised methods. These methods utilize photometric loss to supervise their networks, eliminating the need for optical flow labels. 
Other works, such as \cite{bello2023optical,feng2022disentangling,sun2024dynamo}, have also tried to address consistent depth prediction in a self-supervised manner.
Dynamo-Depth \cite{sun2024dynamo} is one of the few self-supervised methods that can provide pose estimation, but its results are quite poor.
We integrate the aforementioned self-supervised techniques by combining photometric loss with features derived from DINO \cite{caron2021emerging}. This integration supervises the estimation of camera poses, depths, and the recognition and decomposition of dynamic regions within images, enabling our method to significantly outperforms other self-supervised methods in SLAM tasks.

\begin{figure*}[t]
    \centering
    \includegraphics[width=0.9\linewidth]{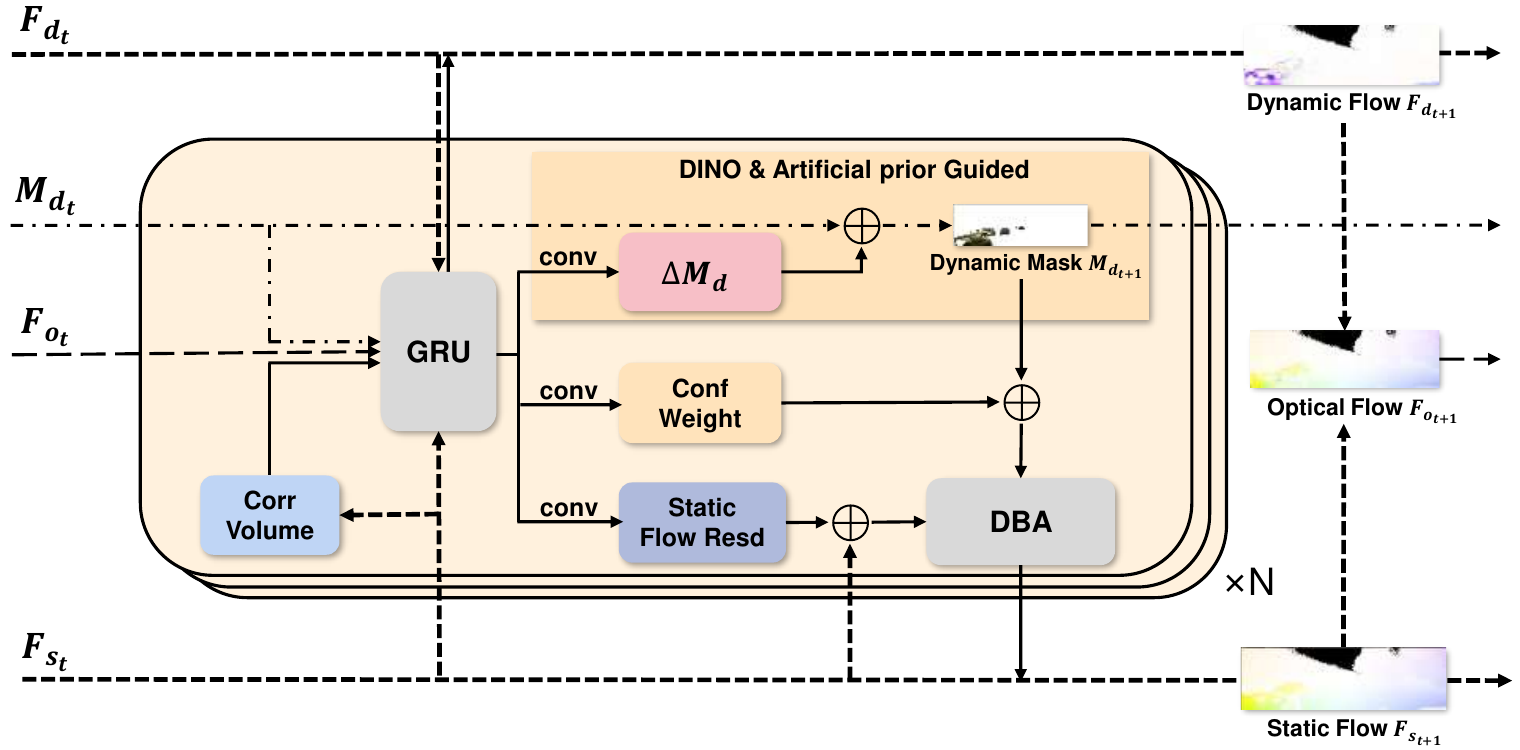}
    \caption{\textbf{Dynamic Update Module.} 
    The correlation feature volume
    with current optical flow $\mathbf{F_{o_t}}$, static flow $\mathbf{F_{s_t}}$ and dynamic mask $\mathbf{M_{d_t}}$
    will be fed to ConvGRU, 
    getting different output splits.
    Adding dynamic mask residual $\mathbf{{\Delta M}_d}$ to the original mask yields new dynamic mask $\mathbf{M_{d_{t+1}}}$, which is simultaneously guided by the DINO prior and artificial prior created through dual-flow representation. Static flow residual $\mathbf{r_s}$ plus original static flow is fed into the dense bundle adjustment (DBA) layer that combines $\mathbf{M_{d_{t+1}}}$ and confidence weight to optimize the depth and pose. Finally, the new static and dynamic flow are summed to get the optical flow.
    }
    \label{fig:update module}
\end{figure*}

\subsection{Scene Motion Decomposition}
Scene motion estimation captures the 3D structure and dynamics of environments, gaining increasing attention in 3D perception. Various methods have been developed using different input data types, such as 3D point clouds~\cite{liu:2019:flownet3d, HPLFlowNet}, stereo images~\cite{huguet2007variational, vogel2013piecewise, zhang20013d}, and RGB-D images~\cite{hornacek2014sphereflow, hadfield2011kinecting, Lv18eccv, quiroga2014dense}. 
EffiScene~\cite{Jiao_2021_CVPR} jointly learns optical flow and motion segmentation for scene flow estimation.
Monocular scene motion estimation remains less explored due to its highly ill-posed nature \cite{brickwedde2019mono}. We focus on monocular videos, framing scene motion estimation as the optical flow~\cite{dosovitskiy2015flownet, ilg2017flownet, sun2018pwc, ranjan2019competitive} problem. While RAFT~\cite{teed2020raft} is efficient in optical flow estimation, it doesn't consider scene motion decomposition.  Based on these works, we decide to decompose optical flow into a static field from camera motion and a dynamic field from object motion, leveraging scene flow properties~\cite{Baur2021ICCV} to provide motion information for each pixel.

\section{Method}
\label{sec:overview}
Fig.~\ref{fig:pipeline} provides an overview of D$^3$FlowSLAM, which processes a sequence of images to produce both the camera pose estimation and the 3D map of the environment. It features an end-to-end differentiable architecture that integrates the strengths of traditional and learning-based approaches. The dual-flow representation enables robust handling of challenging scenarios, including dynamic scenes. Additionally, we design a self-supervised scheme to train our model without labels.


\subsection{Preliminaries: DROID-SLAM}
DROID-SLAM~\cite{teed2021droid} is a deep dense SLAM method. It processes a sequence of images $\{\mathbf{I}_t\}_{t=0}^N$, using a image encoder to extract correlation features from the images. 
It maintains two key state variables for each image $t$: the camera pose $\mathbf{G}_t \in SE(3)$ and the inverse depth $\mathbf{d}_t \in \mathbb{R}_{+}^{H\times W}$. These variables are iteratively updated with each new frame.  
It uses the dense correspondence field to compute the static flow residual $\mathbf{p}_{ij} - \mathbf{p}_j$, which is caused by the camera motion.
A frame graph $\mathcal{G} = (\mathcal{V}, \mathcal{E})$ is used to indicate co-visibility between frames, where each node corresponds to an input image. An edge  $(i, j) \in \mathcal{E}$ represents that the images ${I}_i$ and ${I}_j$ The frame graph is built and updated dynamically during training and inference. Specifically, after updating each pose and depth using the static flow residual field, the frame graph is adjusted to incorporate new co-visibility relationships. For detailed information, please refer to their paper or our appendix.

\subsection{Dual-Flow Representation}

The core concept of our approach is the dual-flow representation coupled with a self-supervised training scheme.
Unlike the method in \cite{teed2021droid}, which uses a simple single-flow representation, our approach decomposes optical flow into two components: static flow driven by camera motion, and dynamic flow driven by the movement of dynamic objects, as depicted in Fig.~\ref{fig:update module}.
The optical flow $\mathbf{F_{o}}_t \in \mathbb{R}^{H\times W\times 2}$, static flow $\mathbf{F_{s}}_t \in \mathbb{R}^{H\times W\times 2}$ and dynamic flow $\mathbf{F_{d}}_t \in \mathbb{R}^{H\times W\times 2}$ are a set of vectors, where the static flow plus the dynamic flow equals the optical flow:
\begin{equation}
 \mathbf{F_{o}}_t = \mathbf{F_{s}}_t + \mathbf{F_{d}}_t.
 \label{equ: flow equ}
\end{equation}

Our network operates on a sequence of images $\{\mathbf{I}_t\}_{t=0}^N$.
\gxy{As new frames are processed, our network iteratively updates not only the camera poses $\{\mathbf{G}_t\}_{t=0}^N \in SE(3)$ and inverse depths $\{\mathbf{d}_t\}_{t=0}^N \in \mathbb{R}_{+}^{H\times W}$, but also the dynamic flows $\{\mathbf{F_{d}}_t\}_{t=0}^N\in \mathbb{R}^{H\times W\times 2}$ and the binary dynamic masks $\{\mathbf{M_{d}}_t\}_{t=0}^N \in \mathbb{R}_{+}^{H\times W \times 2 }$.}
We let 0 indicate dynamic while 1 indicate static in $\mathbf{M_{d}}_t$. 
\subsection{Dynamic Update Module}
Fig.~\ref{fig:update module} demonstrates the dynamic update module of our method, which \gxy{contains} a $3\times3$ ConvGRU with a hidden state $\mathbf{h}$.
Different from the normal update module, which only works on the static flow residual, our dynamic update module works on both static and dynamic flow fields, respectively. For dynamic flow field, 
\gxy{we integrate it with the static flow to compute the optical flow. This combined flow is then fed into the flow encoder as a new optimization term for the subsequent iteration.}
During each iteration, the update module generates a pose increment, depth increment, dynamic mask increment, and dynamic flow. 
The pose increment is applied to the current pose through retraction on the SE3 manifold:
\begin{equation}
    \mathbf{G}^{(k+1)} = \Exp(\Delta \boldsymbol \xi^{(k)}) \circ \mathbf{G}^{(k)}.
\end{equation}
While the depth and the dynamic mask increment are added to the current depth and dynamic mask, respectively:
\begin{equation}
    \Xi^{(k+1)} = \Delta \Xi^{(k)} + \Xi^{(k)}, \, \Xi \in \{\mathbf{d}, \mathbf{M_d}\}.
\end{equation}
And $\mathbf{F_d}^{(k+1)}$ is directly assigned a new value in each iteration. 
With the updated static flow $\mathbf{F_{s}}^{(k+1)}$ transformed from $\mathbf{G}^{(k+1)}$ and $\mathbf{d}^{(k+1)}$, the final optical flow can be computed using Eq.~\ref{equ: flow equ}.

In summary, our dynamic update module produces a sequence of poses, depths, dynamic masks, dynamic flows, and complete optical flows with the expectation of converting to an optimal point, such as
$\{\mathbf{G}^{(k)}\} \rightarrow \mathbf{G}^*$, $\{\mathbf{d}^{(k)}\} \rightarrow \mathbf{d}^*$, $\{\mathbf{M_{d}}^{(k)}\} \rightarrow \mathbf{M_{d}}^*$, $\{\mathbf{F_{d}}^{(k)}\} \rightarrow \mathbf{F_{d}}^*$, $\{\mathbf{F_{o}}^{(k)}\} \rightarrow \mathbf{F_{o}}^*$.

\subsubsection{Dynamic Update Process}
\gxy{Our dynamic update module produces several outputs:}
(1) a  static flow residual field $\mathbf{r_s}_{ij} \in \mathbb{R}^{H \times W \times 2}$, (2) an updated dynamic flow field $\mathbf{F_d}_{ij} \in \mathbb{R}^{H \times W \times 2}$,  (3) a correlation confidence map $\mathbf{w}_{ij} \in \mathbb{R}^{H \times W \times 2}$, (4) an updated dynamic mask increment field $\Delta \mathbf{M_d}_{ij} \in \mathbb{R}^{H \times W \times 2}$. 
\gxy{For initialization, the dynamic mask $\mathbf{M_d}_{ij}$ and the initial dynamic flow $\mathbf{F_d}_{ij}$ are both set to zeros. }
The residual $\mathbf{r_s}_{ij}$ is
used to correct errors in the dense correspondence fields, which can be expressed as $\mathbf{p_s}_{ij}^* = \mathbf{r_s}_{ij} + \mathbf{p_s}_{ij}$. 
The maps predicted by the module are in low resolution. \ywc{To obtain the original resolution maps, we follow the mask upsample method mentioned in~\cite{teed2020raft} to get better upsample results}.

\subsubsection{Dense Bundle Adjustment Layer}
\ywc{After we get the static flow residual field, we optimize the poses and depth maps using a dense bundle adjustment layer \cite{teed2021droid}. Its cost function is defined as follows}:
\begin{equation}
    \mathbf E(\mathbf{G}', \mathbf{d}') = \sum_{(i,j) \in \mathcal{E}} \norm{\mathbf{p_s}_{ij}^* - \Pi_c(\mathbf{G}'_{ij} \circ \Pi_c^{-1}(\mathbf{p}_i, \mathbf{d}'_i)) }_{\Sigma_{ij}}^2
    \label{eqn:objective},
\end{equation}
\begin{equation}
\Sigma_{ij} = \diag \mathbf{w_d}_{ij},
\end{equation}
\begin{equation}
    \mathbf{w_d}_{ij}=\operatorname{sigmoid}(\mathbf{w}_{ij}-(1-\mathbf{M_d}_{ij})\cdot\eta),
\end{equation}
where $\eta$ is set as 10. $\norm{\cdot}_{\Sigma}$ is the Mahalanobis distance, which weights the error term according to the modified confidence $\mathbf{w_d}_{ij}$. Note that we use the dynamic mask $\mathbf{M_d}_{ij}$ here to explicitly mitigate the impact of dynamic regions, ensuring the quality of point correspondences during the BA process. It improves the optimization process in dynamic environments, minimizing the negative impact of these points.
During BA optimization, we use Gauss-Newton algorithm to solve the linear system.

\subsection{Loss Function}

\subsubsection{Geometry Photometric Loss}
To supervise the geometric predictions in a self-supervised scheme, we introduce a photometric reprojection loss \cite{2020Digging} to guide the network's optimization. 
Given the predicted pose $\mathbf{G_{ij}}$ and the predicted depth $\mathbf{\hat{d}}_i$, we can get the corresponding coordinates of pixels in image $I_i$ in image $I_j$. We then use bi-linear sampling to sample the image $I_j$, getting a new-sampled image $I_{j \rightarrow i}$:
\begin{equation}
    I_{j \rightarrow i}=I_{j}\left\langle\Pi_c(\mathbf{G}_{ij} \circ \Pi_c^{-1}(\mathbf{p}_i, \mathbf{\hat{{d}}}_i))\right\rangle.
\end{equation}
Then we can use the photometric loss on source image $I_i$ and new image $I_{j \rightarrow i}$:
\begin{equation}
    \mathcal{L}_{\text {geo\_ph}}=\frac{1}{N}\sum_{ij} p e\left(I_{i}, I_{j \rightarrow i}\right).
\end{equation}
We leverage $L_1$ loss and $\operatorname{SSIM}$ \cite{2004Image} loss to form our geometry photometric loss with $\alpha=0.85$:
\begin{equation}
    p e\left(I_{a}, I_{b}\right)=\frac{\alpha}{2}\left(1-\operatorname{SSIM}\left(I_{a}, I_{b}\right)\right)+(1-\alpha)\left\|I_{a}-I_{b}\right\|_{1}.
\end{equation}

\subsubsection{DINO Mask Guidance for Geometry}
In our self-supervised training scheme, directly using the dual-flow representation in the photometric loss may lead to pixel mismatches due to object motion.
Relying solely on optical flow warping may cause the convergence of two different flows, compromising the network's ability to decompose the scene. Therefore, a robust prior is needed during training to help the network achieve an initial scene decomposition.
\gxy{Assuming that most dynamic objects in the scene are foreground entities, we use the foreground regions as a prior. 
}
We adopt the method proposed by~\cite{amir2021deep}, utilizing the high-quality and generalizable DINO features \cite{caron2021emerging} to cluster the foreground parts $\mathbf{M_{d_i}^{dino}}$ of the scene images. This prior is used to filter out potential dynamic pixel matches, enhancing our model's training process.
So the final geometry photometric loss function is:
\begin{equation}
     \mathcal{L}_{\text {geo\_ph}}=\frac{1}{N^{'}}\sum_{ij} p e\left(I_{i}, I_{j \rightarrow i}\right) \cdot \mathbf{M_{d_i}^{dino}},
\end{equation}
where $N^{'}$ means the count of pixels whose $\mathbf{M_{d_i}^{dino}}$ value is 1. 
Tab.~\ref{table:vkitti ablation study} shows the guidance strategy helps filter out the ambiguous matches, obtaining better results.

\subsubsection{Optical Flow Photometric Loss}
The geometric photometric loss is used to supervise the static flow caused by camera motion. For the remaining part, we introduce an optical flow photometric loss to supervise the complete scene motion, including both camera and object motion.
\gxy{Through the update module, the optical flow results $\mathbf{F_o}_{ij}$ are derived by combining static and dynamic flows.}
Similar to $\mathcal{L}_{\text {geo\_ph}}$, we use $\mathbf{F_o}_{ij}$ to generate corresponding coordinates between images:
\begin{equation}
    I_{j \rightarrow i}=I_{j}\left\langle \mathbf{F_o}_{ij}+ \mathbf{p_{ij}}\right\rangle.
\end{equation}
Using bi-linear sampling to sample from the source image, we evaluate their photometric errors:
\begin{equation}
    \mathcal{L}_{\text {flow\_ph}}=\sum_{ij} p e\left(I_{i}, I_{j \rightarrow i}\right),
\end{equation}
where the $pe$ function here is just $L_1$ loss:
\begin{equation}
    p e\left(I_{a}, I_{b}\right)=\left\|I_{a}-I_{b}\right\|_{1}.
\end{equation}

\subsubsection{DINO Guided Mask Loss}
The mask $\mathbf{M_{d_i}^{dino}}$ serves as a reliable prior for identifying potential dynamic regions during the early stages, when the network’s predictions may lack precision.
\gxy{Consequently, } we directly use it to supervise our predicted masks using a cross-entropy classification loss~\cite{jadon2020survey}:
\begin{align}
    \mathcal{L}_{\text{dino\_mask}} = &-\frac{1}{\left|\mathcal N\right|} \sum_{\mathbf{p}_{i} \in \mathcal N} \mathbf{M_{d_i}^{dino}} \log \mathbf{\hat M}_i + \nonumber\\
    &\left(1-\mathbf{M_{d_i}^{dino}}\right) \log \left(1-\mathbf{\hat M}_i\right),
\end{align}
where $\mathbf{\hat M}_i$ represents the predicted mask.

\subsubsection{Artificial Mask Loss}
The mask $\mathbf{M_{d_i}^{dino}}$ provides a potential representation of dynamic regions, though it is not yet accurate enough. To enhance this, we adopt a method similar to \cite{Baur2021ICCV}, using the dual-flow representation to artificially construct an additional prior mask that serves as both a supplement and a form of regularization.
With the camera poses, depths, and optical flows we have already obtained, we can infer the target coordinate of pixel $p$ using the following equations:
\begin{equation}
    \mathbf{p}_{\text {cam}} =\Pi_c(\mathbf{G}_{ij} \circ \Pi_c^{-1}(\mathbf{p}_i, \mathbf{\hat{{d}}}_i)), \\
    \mathbf{p}_{\text {flow}} =\mathbf{p}_i+ \mathbf{\hat F_o{}}_{ij},
\end{equation}
where $\mathbf{p}_{\text{cam}}$ is the target coordinate calculated by projection, $\mathbf{p}_{\text{flow}}$ is the target coordinate calculated by optical flow.
We then use the difference between these two coordinates to create our artificial mask prior, where sufficiently large difference indicates dynamic regions:
\begin{equation}
    \mathbf{M^{art}_{i}}=[\left\|\mathbf{p}_{\text {cam}}-\mathbf{p}_{\text {flow}}\right\|_{2}\le \mu],
\end{equation}
where $\mu$ is set as 0.5. This artificial mask prior has the same format as $\mathbf{M_{d_i}^{dino}}$, so our final loss function should be:
\begin{align}
\mathcal{L}_{\text{art\_mask}} = &-\frac{1}{\left|\mathcal N\right|} \sum_{\mathbf{p}_{i} \in \mathcal N} \mathbf{M^{art}_{i}} \log \mathbf{\hat M}_i + \nonumber\\ 
    &\left(1-\mathbf{M^{art}_{i}}\right) \log \left(1-\mathbf{\hat M}_i\right).
\end{align}

\begin{figure}[t]
    \centering
    \includegraphics[width=1\linewidth]{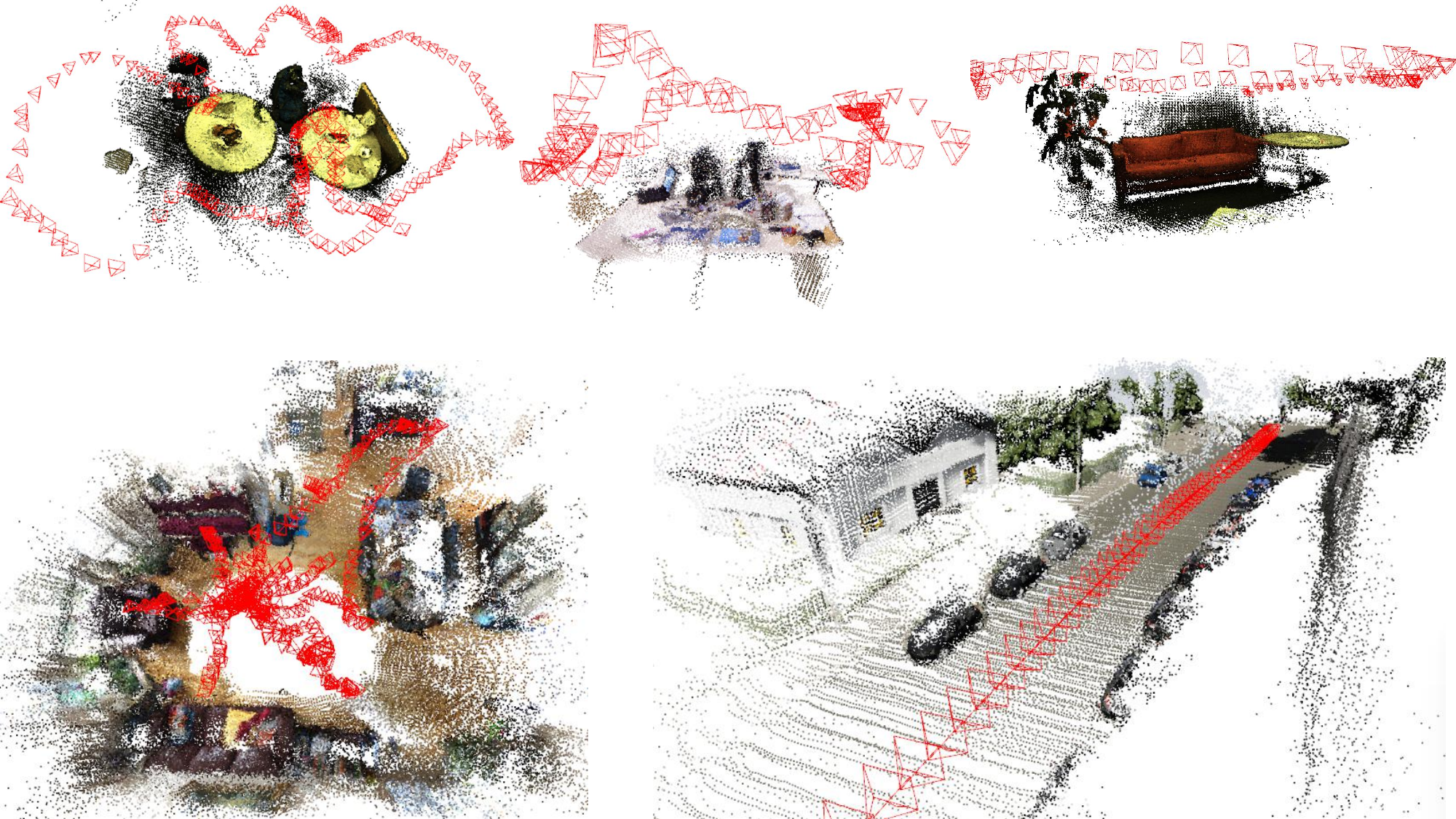}
    \caption{\textbf{Reconstruction visualizations of our method}. Our method can generalize to different datasets.}
    \label{fig:slam reconstruction}
\end{figure}

\begin{table*}[t]
    \centering
    \caption{\textbf{Dynamic SLAM Results on KITTI(K) \& Virtual KITTI2(VK) Datasets with Metric: ATE[m].} 
    Bold stands for best results.
    * means the results are generated by running the official pretrained model in our environment to ensure evaluation consistency. X means the system has failed here. - means lack of results.} 
    \footnotesize
    \resizebox{1\linewidth}{!}
    {
    {%
    \begin{tabular}{cl| c c c c c c c |c}
        & Method & K09 & K10 & VK01 & VK02 & VK06 & VK18 & VK20 & Avg \\
        \midrule
        \multirow{3}{*}{\rotatebox[origin=c]{0}{Sup}}
        &DynaSLAM* \cite{bescos2018dynaslam} & {18.910} & \textbf{7.519} & 27.830 & X & X & X & \textbf{2.807} & - \\
        &DROID-SLAM* \cite{teed2021droid} & 47.053 & 11.001 & {2.259} & \textbf{0.049} & {0.136} & {1.170} & 6.998 & 9.816 \\
        &PVO \cite{ye2023pvo} & \textbf{14.650} &  8.660 &  \textbf{0.369} & {0.055} &  \textbf{0.113} &  \textbf{0.822} &  {3.079} & \textbf{3.964}\\
        \midrule
        \multirow{2}{*}{\rotatebox[origin=c]{0}{Trad}}
        &DSO \cite{engel2017direct} & {28.100} & 24.000 & - & - & - & - & - & -\\
        &ORB-SLAM \cite{mur2015orb} & \textbf{6.620} & \textbf{8.800} & - & - & - & - & - & -\\
        \midrule
        \multirow{2}{*}{\rotatebox[origin=c]{0}{S-Sup}}
        & Dynamo-Depth \cite{sun2024dynamo} & 130.536 & 62.417 & 33.516 & 1.232 & 2.035 & 10.761 &  22.058 & 37.508 \\
        &Ours & \textbf{10.608} & \textbf{3.984} & \textbf{1.151} & \textbf{0.109} & \textbf{0.093} & \textbf{0.750} & \textbf{3.269} & \textbf{2.852} \\
        \bottomrule
    \end{tabular}
    }
    }
    \label{tab:Kitti}
\end{table*}

\begin{table*}[t]
    \centering
    \caption{\textbf{Dynamic SLAM Results on TUM Dynamic Sequences with Metric: ATE[m].} 
    The rest references are DVO SLAM \cite{kerl2013robust}, ORB-SLAM2 \cite{orbslam2} and PointCorr \cite{dai2020rgb}.
    }
    
    \label{tab:tum_dy}
    \footnotesize
    \resizebox{1.0\linewidth}{!}
    {
    \begin{tabular}{cl|ccccc|cccc|c}
        & & \multicolumn{5}{c|}{slightly dynamic} & \multicolumn{4}{c|}{highly dynamic} & \\
        & Method & fr2/d-person & fr3/s-static & fr3/s-xyz & fr3/s-rpy & fr3/s-half & fr3/w-static & fr3/w-xyz & fr3/w-rpy & fr3/w-half & Avg \\
        \midrule
        \multirow{3}{*}{Sup} & DROID-SLAM & 0.017 & {0.007} & {0.016} & {0.029} & \textbf{0.022} & {0.016} & {0.019} & {0.059} & 0.312 & {0.055} \\
        & PVO & 0.013 & \textbf{0.006} & 0.014 & 0.027 & \textbf{0.022} & \textbf{0.007} & \textbf{0.018} & \textbf{0.056} & 0.221 & 0.043 \\
        & Point-Corr & \textbf{0.008} & 0.010 & \textbf{0.009} & \textbf{0.023} & 0.024 & {0.011} & 0.087 & 0.161 & \textbf{0.035} & \textbf{0.041} \\
        \midrule
        \multirow{2}{*}{Trad} & DVO-SLAM & 0.104 & 0.012 & 0.242 & 0.176 & 0.220 & 0.752 & 1.383 & 1.292 & 1.014 & 0.688 \\
        & ORB-SLAM2 & \textbf{0.006} & \textbf{0.008} & \textbf{0.010} & \textbf{0.025} & \textbf{0.025} & \textbf{0.408} & \textbf{0.722} & \textbf{0.805} & \textbf{0.723} & \textbf{0.303} \\
        \midrule
        \multirow{2}{*}{S-Sup} & Dynamo-Depth & 1.173 & 0.015 & 0.299 & 0.051 & 0.379 & 0.018 & 0.283 & 0.153 & 0.458 & 0.314 \\
        & Ours & \textbf{0.069} & \textbf{0.007} & \textbf{0.018} & \textbf{0.039} & \textbf{0.117} & \textbf{0.008} & \textbf{0.112} & \textbf{0.143} & \textbf{0.114} & \textbf{0.070} \\
        \bottomrule
    \end{tabular}
    }
\end{table*}

\begin{table*}[t]
    \centering
    \caption{
        \textbf{Monocular SLAM Results on TartanAir Monocular Benchmark with Metric: ATE[m]}. 
    }
    \footnotesize
    \resizebox{1.0\linewidth}{!}{%
        \begin{tabular}{cl|cccccccc | c}
            &Method & MH000 & MH001 & MH002 & MH003 & MH004 & MH005 & MH006 & MH007 & Avg \\
            \toprule
            \multirow{3}{*}{\rotatebox[origin=c]{0}{Sup}} 
            &DeepV2D~\cite{deepv2d} & 6.15 & 2.12 & 4.54 & 3.89 & 2.71 & 11.55 & 5.53 & 3.76 & 5.03 \\
            &TartanVO~\cite{wang2021tartanvo} & 4.88 & \textbf{0.26} & 2.00 & {0.94} & \textbf{1.07} & 3.19 & 1.00 & 2.04 & 1.92 \\
            &DROID-SLAM*~\cite{teed2021droid} & \textbf{0.04} & 0.69 & \textbf{0.03} & \textbf{0.02} & 3.73 & \textbf{1.29} & \textbf{0.38} & \textbf{0.07} & \textbf{0.78} \\
            \midrule
            \multirow{2}{*}{\rotatebox[origin=c]{0}{Trad}} 
            &ORB-SLAM3~\cite{orbslam3} & 15.44 & 2.92 & 13.51 & 8.18 & \textbf{2.59} & 21.91 & 11.70 & 25.88 & \textbf{12.77} \\
            &DSO~\cite{dso} & \textbf{9.92} & \textbf{0.35} & \textbf{7.96} & \textbf{3.46} & - & \textbf{12.58} & \textbf{8.42} & \textbf{7.50} & - \\
            \midrule
            \multirow{2}{*}{\rotatebox[origin=c]{0}{S-Sup}} 
            &Dynamo-Depth \cite{sun2024dynamo} & 33.36 & 7.67 & 15.89 & 10.13 & 9.39 & 19.46 & 16.94 & 15.08 & 15.99 \\
            &Ours & \textbf{0.83} & \textbf{0.11} & \textbf{0.19} & \textbf{0.36} & \textbf{2.41} & \textbf{1.62} & \textbf{0.19} & \textbf{0.58} & \textbf{0.79} \\
            \bottomrule
        \end{tabular}
        \label{table:TartanAir mono}
    }
\end{table*}

\begin{table*}[t]
    \caption{
        \textbf{Monocular SLAM Results on EuRoC Dataset with Metric: ATE[m]}. 
    }
    \centering
    \footnotesize
    \resizebox{1.0\linewidth}{!}{%
        \begin{tabular}{cl| ccccc | ccc | ccc | c}
            & Method & MH01 & MH02 & MH03 & MH04 & MH05 & V101 & V102 & V103 & V201 & V202 & V203 & Avg \\
            \toprule
            \multirow{5}{*}{\rotatebox[origin=c]{0}{Sup}}
            & DeepFactors~\cite{deepfactors} & 1.587 & 1.479 & 3.139 & 5.331 & 4.002 & 1.520 & 0.679 & 0.900 & 0.876 & 1.905 & 1.021 & 2.040 \\
            & DeepV2D~\cite{deepv2d}
            & 0.739 & 1.144 & 0.752 & 1.492 & 1.567 & 0.981 & 0.801 & 1.570 & 0.290 & 2.202 & 2.743 & 1.298 \\
            & TartanVO~\cite{wang2021tartanvo}
            & 0.639 & 0.325 & 0.550 & 1.153 & 1.021 & 0.447 & 0.389 & 0.622 & 0.433 & 0.749 & 1.152 & 0.680 \\
            & D3VO + DSO~\cite{d3vo}
            & - & - & 0.08 \ \ & - & 0.09 \ \ & - & - & 0.11 \ \ & - & 0.05 \ \ & {0.19} \ \ & - \\
            & DROID-SLAM*~\cite{teed2021droid} & \textbf{0.013} & \textbf{0.014} & \textbf{0.022} & \textbf{0.043} & \textbf{0.043} & \textbf{0.037} & \textbf{0.012} & \textbf{0.020} & \textbf{0.017} & \textbf{0.013} & \textbf{0.014} & \textbf{0.022} \\
            \midrule
            \multirow{4}{*}{\rotatebox[origin=c]{0}{Trad}}
            & DSO~\cite{dso}
            & 0.046 & 0.046 & 0.172 & 3.810 & 0.110 & 0.089 & 0.107 & 0.903 & 0.044 & 0.132 & 1.152 & 0.601 \\
            & SVO~\cite{svo}
            & 0.100 & 0.120 & 0.410 & 0.430 & 0.300 & 0.070 & 0.210 & X & 0.110 & 0.110 & 1.080 & - \\
            & DSM~\cite{dsm} & {0.039} & {0.036} & {0.055} & \textbf{0.057} & \textbf{0.067} & 0.095 & {0.059} & {0.076} & 0.056 & {0.057} & 0.784 & \textbf{0.126} \\
            & ORB-SLAM3~\cite{orbslam3} & \textbf{0.016} & \textbf{0.027} & \textbf{0.028} & 0.138 & {0.072} & \textbf{0.033} & \textbf{0.015} & \textbf{0.033} & \textbf{0.023} & \textbf{0.029} & X & - \\
            \midrule
            \multirow{2}{*}{\rotatebox[origin=c]{0}{S-Sup}} & Dynamo-Depth \cite{sun2024dynamo} & 4.257 & 3.773 & 3.430 & 6.298 & 6.460 & 1.762 & 1.746 & 1.480 & 2.017 & 2.054 & 1.906 & 3.198 \\
            & Ours & \textbf{0.164} & \textbf{0.184} & \textbf{0.210} & \textbf{0.243} & \textbf{0.775} & \textbf{0.058} & \textbf{0.161} & \textbf{1.360} & \textbf{0.041} & \textbf{1.751} & \textbf{1.228} & \textbf{0.561} \\
            \bottomrule
        \end{tabular}
    }
    \label{table:EurocMono}
\end{table*}

\subsubsection{Complete Loss Function}
Our complete loss function is:
\begin{equation}
    \mathcal{L}_{\text{self-sup}}=\lambda_1\mathcal{L}_{\text {geo\_ph}}+\lambda_2\mathcal{L}_{\text{flow\_ph}}+\lambda_3\mathcal{L}_{\text{dino\_mask}} +\lambda_4\mathcal{L}_{\text{art\_mask}},
\end{equation}
where $\lambda_1=100$, $\lambda_2=5$, $\lambda_3=0.05$, and $\lambda_4=0.05$. Since the network has several update iterations, we use $\gamma=0.9$ to apply the loss to the output of each iteration with exponentially increasing weights.


\subsection{Implementation Details}

We implement D$^3$FlowSLAM in PyTorch and use the LieTorch extension~\cite{teed2021tangent} to perform back-propagation in the tangent space of all group elements. In the ablation study, we train our model on the Virtual KITTI2 dataset~\cite{cabon2020virtual} with 2 RTX-3090 GPUs for 80,000 steps, which takes about 2.5 days. 
In the full training, we use the TartanAir dataset~\cite{tartanair2020iros} and the PointOdyssey dataset~\cite{zheng2023pointodyssey} together.
It takes 9 days for 300k steps on 4 RTX-3090 GPUs. We use Adam \cite{kingma2014adam} optimizer with a learning rate of 0.00025. 

During training, we use a 6-frame video sequence as one batch data.
In temporal dimension, we randomly select frames to add to the sequence with a specific step size. Both the step size and the overall span of the sequence are constrained. For the first frame pair in each batch, we use the 2D-2D epipolar constraint function provided by OpenCV~\cite{bradski2000opencv} to directly initialize the pose transformation between the first and second frame images. This initialization step enhances training effectiveness and accelerates network convergence. 
During evaluation, the priors provided by the DINO model are no longer needed. We keep the same system settings in \cite{teed2021droid}, achieving 10 fps at a resolution $240 \times 808$ and 15 fps at $240 \times 320$. Other details can be found in our appendix.

\section{Experiments}


We train our model in a \textbf{self-supervised scheme} and test our model on different datasets.
We use absolute trajectory error (ATE)~\cite{tumrgbd} to evaluate the accuracy of the estimated camera trajectories. 
We compare our method with traditional methods (Trad), learning-based supervised methods (Sup), and the latest \textbf{self-supervised methods} (S-Sup) \textbf{similar to ours}.
Specifically, we choose Dynamo-Depth \cite{sun2024dynamo} because it is the best method among existing self-supervised approaches for providing pose estimation in dynamic scenes.
\subsection{Datasets}
We validate the effectiveness of our method in highly dynamic scenes like Virtual KITTI2~\cite{cabon2020virtual} in the ablation study. 
Virtual KITTI2 is derived from the KITTI benchmark \cite{geiger2013vision} and consists of 5 sequences augmented with various weather conditions. We use the \textit{clone} split for training and the \textit{15-degree} split for testing. 

To get our final weight, we use larger datasets like TartanAir~\cite{tartanair2020iros} and PointOdyssey~\cite{zheng2023pointodyssey} for training.
TartanAir is a large synthetic static dataset that includes a rich variety of simulated images from both indoor and outdoor scenes. 
As a supplement, we include the PointOdyssey dataset, which is another large synthetic dataset containing a wealth of dynamic scene data.

We test our method on various dynamic datasets such as Virtual KITTI2, KITTI, dynamic sequences from TUM-RGBD \cite{schubert2018tum}, and traditional SLAM datasets including static scenes from TUM-RGBD, EuRoC~\cite{burri2016euroc}, and TartanAir-Test \cite{tartanair2020iros}.
The visualizations in Fig.~\ref{fig:slam reconstruction} show that our model runs well in different datasets. For the final evaluation on different datasets, we use the self-supervised nature to perform simple finetuning on our final weight and Dynamo-Depth's official weight with specific image sequences from these datasets.
Details can be found in our appendix.

\begin{figure}[t]
  \centering
  \includegraphics[width=1.0\linewidth]{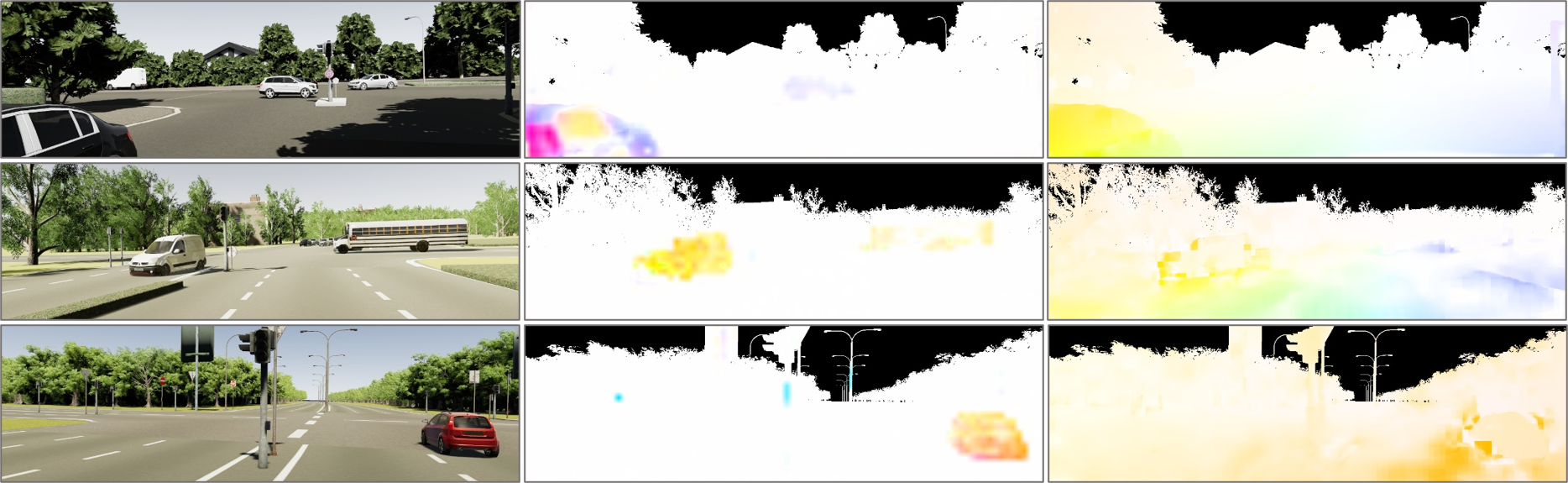}
  \caption{\textbf{Flow decomposition visualization.} From left to right: Input image, dynamic flow, and static flow.}
  \label{fig:flow decomp}
\end{figure}

\subsection{Dynamic SLAM}

\begin{table}[t]
    \centering
    \caption{\textbf{Ablation Study 
    on Virtual KITTI2(VK) Dataset.} SF means single flow, DF means dual flow, and DINO means DINO Guidance. 
    }
    \footnotesize
    \resizebox{1.0\linewidth}{!}{%
        \begin{tabular}{l|c c c c c}
            Method & VK01 & VK02 & VK06 & VK18 & VK20 \\
            \toprule
            DROID-SLAM (Sup) & {1.091} & \textbf{0.025} & 0.113 & {1.156} & {8.285} \\
            Ours (SF) & 3.281 & 0.194 & \textbf{0.093} & 1.183 & 10.259\\
            Ours (DF) & 4.045 & 0.185 & {0.095} & 2.141 & 8.782 \\
            \midrule
            \textbf{Ours (DF, DINO)} & \textbf{0.793} & {0.029} & 0.121 & \textbf{0.382} & \textbf{3.083} \\
            \bottomrule
        \end{tabular}
        \label{table:vkitti ablation study}
    }
\end{table}

In dynamic settings, we test on sequences \textit{09} and \textit{10} from the KITTI (K) dataset and all sequences from the Virtual KITTI2 (VK) dataset. 
ATE results are shown in Tab.~\ref{tab:Kitti}. We also test on TUM-RGBD dynamic sequences, as shown in Tab.~\ref{tab:tum_dy}. Our method significantly outperforms the best existing self-supervised methods, especially in challenging dynamic scenes like K and VK. While Dynamo-Depth's estimation results are subpar in these scenarios, our method still delivers accurate estimations. Compared to traditional or supervised methods, 
our method also achieves comparable or better results. We achieve an average ATE of 2.852m on VK+K and 0.07m on TUM-RGBD dynamic,
which confirm the effectiveness of our method in dynamic scenes.



\subsection{Monocular SLAM}

In monocular settings, we test our method on TartanAir-Test, EuRoC, and TUM-RGBD dataset. Results for TUM-RGBD can be found in our appendix.
Tab.~\ref{table:TartanAir mono} and Tab.~\ref{table:EurocMono} show that in static scenes, our method also substantially outperforms other self-supervised methods. Compared to existing supervised methods and traditional methods, our results remain competitive. 
Specifically, we achieve an average ATE of 0.79m on TartanAir-Test, 0.561m on EuRoC in the monocular setting, and 0.127m on TUM-RGBD static sequences.

\subsection{Ablation Study}

\begin{figure}[t]
  \centering
  \includegraphics[width=1.0\linewidth]{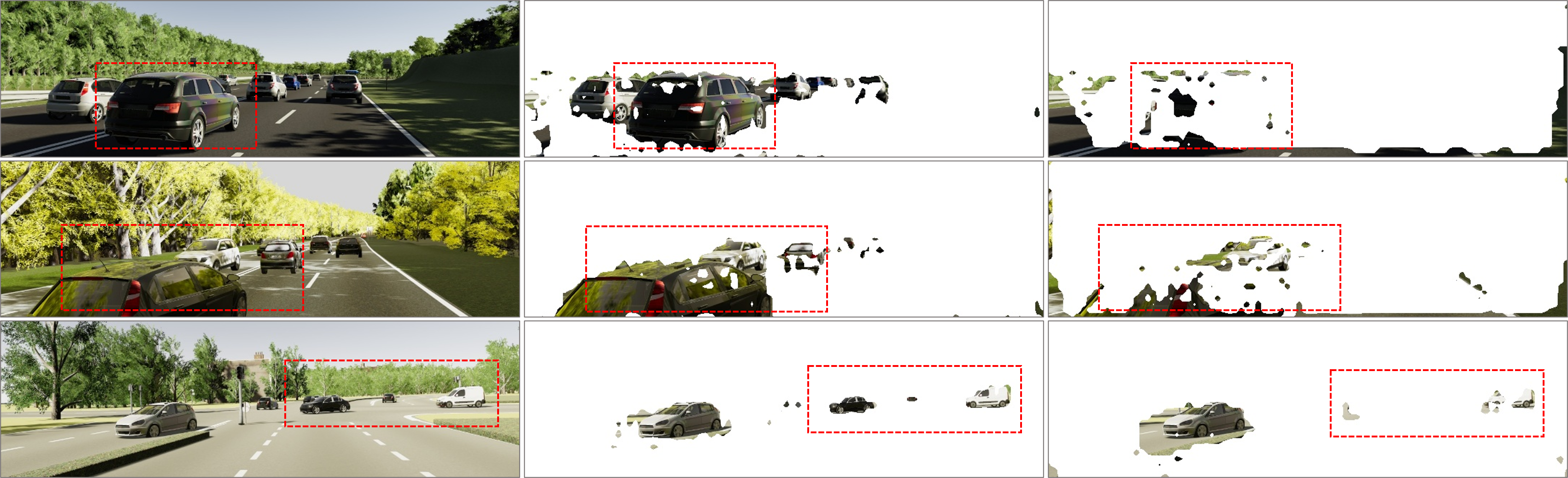}
  \caption{\textbf{Ablation for DINO Guidance.} From left to right: Input image, prediction with DINO, and prediction without DINO. Colorful masks indicate dynamic parts.}
  \label{fig:motion segmentation}
\end{figure}

We conduct an ablation study to verify the effectiveness of our dual-flow representation and DINO Guidance. 
We use the Virtual KITTI2 dataset for both training and testing in the ablation study. The experimental results are detailed in Tab.~\ref{table:vkitti ablation study}. Specifically, SF means only using $\mathcal{L}_{\text{geo\_ph}}$, DF means using $\mathcal{L}_{\text{flow\_ph}}$ and $\mathcal{L}_{\text{geo\_ph}}$, DINO means adding $\mathcal{L}_{\text{dino\_mask}}$. The results show that the dual-flow (DF) representation outperforms the rough single-flow (SF) approach. Additionally, DINO Guidance enhances the pose estimation accuracy and greatly improves our model's ability to decompose dynamic scenes. It provides a stronger prior and reduces the impact of pixel matching errors. Fig.~\ref{fig:motion segmentation} offers a more intuitive demonstration of the improvements brought by DINO Guidance.
Additionally, compared to the results in Tab.~\ref{tab:Kitti}, our model achieves comparable or better results after full training and finetuning.
Additionally, the results from Tab.~\ref{tab:Kitti} and \ref{table:vkitti ablation study} demonstrate that our model, when trained on large datasets and then finetuned, achieves performance comparable to training directly on specific datasets.

\section{Conclusions}
In D$^3$FlowSLAM, we propose a dual-flow representation that decomposes optical flow into static flow and dynamic flow. Based on this, our dynamic update module simultaneously updates these two flow fields, enabling geometric information retrieval and scene decomposition in both static and dynamic environments. Additionally, we design a self-supervised scheme that enables label-free training and significantly outperforms existing self-supervised methods. 
Our limitations include the need for a GPU and a limited fps. Additionally, hyperparameters must be adjusted for each dataset to ensure effective convergence of the self-supervised training. In future work, we plan to enhance the computational efficiency of our model and its training scheme to improve adaptability and robustness, while also further increasing the model's prediction accuracy.

\bibliography{main}

\newpage
\clearpage
\appendix

\setcounter{page}{11}

\twocolumn[
    \centering
    \Large
    \textbf{D$^3$FlowSLAM: Self-Supervised Dynamic SLAM with \\Flow Motion Decomposition and DINO Guidanc} \\
    \vspace{0.5em}Supplementary Material \\
    \vspace{1.0em}
] 

\setcounter{table}{0}
\setcounter{figure}{0}
\setcounter{equation}{0}
\renewcommand{\thetable}{\thesection\arabic{table}}
\renewcommand{\thefigure}{\thesection\arabic{figure}}
\renewcommand{\theequation}{\thesection\arabic{equation}}

In this appendix, we will discuss some details not elaborated in the main paper, along with additional experimental results and visualizations.

\section{Preliminaries: DROID-SLAM}
DROID-SLAM~\cite{teed2021droid} operates on a sequence of images $\{\mathbf{I}_t\}_{t=0}^N$, and maintains two state variables: camera pose $\mathbf{G}_t \in SE(3)$ and inverse depth $\mathbf{d}_t \in \mathbb{R}_{+}^{H\times W}$ for each image $t$, which are iteratively updated as new frames are processed. A frame graph $\mathcal{G} = (\mathcal{V}, \mathcal{E})$ is used to represent co-visibility between frames, where the nodes correspond to input images, and an edge $(i, j) \in \mathcal{E}$ indicates that images ${I}_i$ and ${I}_j$ have overlapping views.

\subsection{Feature Extraction and Correlation}
Following RAFT \cite{teed2020raft}, the input images are first processed by the feature extraction module, after which the relationship between the two images is computed.

\noindent\textbf{Feature Extraction.}
First, an image encoder with six residual blocks and three downsampling layers processes each image, producing a dense feature map at 1/8 of the original resolution. These feature maps from input image pairs are then used for constructing the correlation volume.

\noindent\textbf{Correlation Pyramid.}
\ywc{For each edge $(i, j) \in \mathcal{E}$ in the frame graph $\mathcal{G}$, DROID-SLAM computes the correlation volume $\mathbf{C}^{ij}$ as the dot product of feature vector pairs taken from $\mathbf{f}_\theta(I_i)_{u_i v_i}$ and $\mathbf{f}_\theta(I_j)_{u_j v_j}$ }:

\begin{equation}
    C_{u_i v_i u_j v_j}^{ij} = \langle \mathbf{f}_\theta(I_i)_{u_i v_i},\  \mathbf{f}_\theta(I_j)_{u_j v_j} \rangle,
\end{equation}
where $u_i,v_i,u_j,v_j$ represent the pixel coordinates for image $I_i, I_j$ respectively, and $< . >$ stands for the dot product.
The last two dimensions of the correlation volume are fed to the average pooling layers with four different kernel sizes (1,2,4,8), forming a 4-level correlation pyramid~\cite{teed2020raft}.

\noindent\textbf{Correlation Lookup.}
RAFT defines a correlation lookup operator that uses a coordinate grid with radius $r$ to index the correlation volume $L_r: \mathbb{R}^{H\times W\times H\times W} \times \mathbb{R}^{H\times W \times 2} \mapsto \mathbb{R}^{H\times W \times (r+1)^2}$. This operator takes an $H\times W$ grid of optical flow coordinates as input and retrieves values from the correlation volume using bi-linear interpolation. These values are then concatenated to compute the final feature vector. The lookup function is applied to each correlation volume in the pyramid.


\begin{figure}[t]
  \centering
  \includegraphics[width=\linewidth]{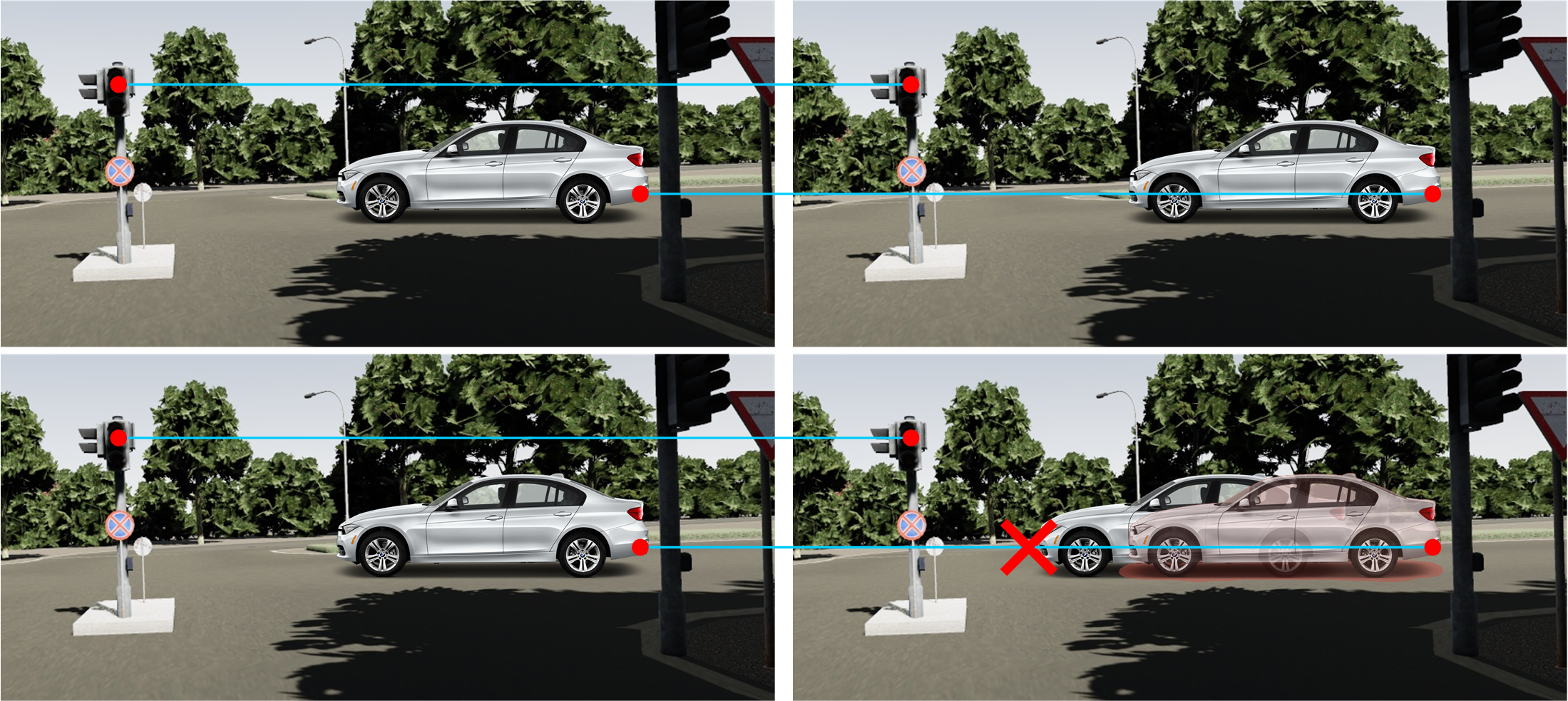}
  \caption{\textbf{Dynamic DINO Guidance Illustration}. If there are dynamic objects in the scene, directly matching pixels using static flow will cause erroneous results, making photometric loss invalid. Using the DINO mask as a prior, we can filter out most foreground parts, which are possible invalid pixels (e.g. the foreground car, pink mask) in geometry photometric loss, which is shown in the picture.}
  \label{fig:maskagg}
\end{figure}

\begin{table*}[t]
    \caption{
        \textbf{Monocular SLAM Results on TUM-RGBD Dataset with Metric: ATE[m]}.
        Bold stands for best results. * means the results are generated by running the official pretrained model in our environment to ensure evaluation consistency. X means the system has failed here. - means lack of results.
    }
    \centering
    \footnotesize
    \resizebox{1\linewidth}{!}{%
        \begin{tabular}{cl|ccccccccc | c}
            & Method & 360 & desk & desk2 & floor & plant & room & rpy & teddy & xyz & Avg \\
            \toprule
            \multirow{5}{*}{\rotatebox[origin=c]{0}{Sup}}
            & DeepV2D~\cite{deepv2d} & 0.243 & 0.166 & 0.379 & 1.653 & 0.203 & 0.246 & 0.105 & 0.316 & 0.064 & 0.375 \\
            & DeepFactors~\cite{deepfactors} & 0.159 & 0.170 & 0.253 & {0.169} & 0.305 & 0.364 & {0.043} & 0.601 & {0.035} & 0.233\\
            & TartanVO~\cite{wang2021tartanvo} & 0.178 & 0.125 & 0.122 & 0.349 & 0.297 & 0.333 & 0.049 & 0.339 & 0.062 & 0.206\\
            & DeepTAM~\cite{deeptam} & \textbf{0.111} & 0.053 & {0.103} & 0.206 & 0.064 & {0.239} & 0.093 & {0.144} & 0.036 & {0.116} \\
            & DROID-SLAM*~\cite{teed2021droid} & \textbf{0.111} & \textbf{0.018} & \textbf{0.042} & \textbf{0.021} & \textbf{0.016} & \textbf{0.049} & \textbf{0.026} & \textbf{0.048} & \textbf{0.012} & \textbf{0.038} \\
            \midrule
            \multirow{1}{*}{\rotatebox[origin=c]{0}{Trad}} 
            & ORB-SLAM3~\cite{orbslam3} & X & {0.017} & 0.210 & X & {0.034} & X & X & X & {0.009} & - \\
            \midrule
            \multirow{2}{*}{\rotatebox[origin=c]{0}{S-Sup}} 
            & Dynamo-Depth \cite{sun2024dynamo} & \textbf{0.177} & 0.838 & 0.930 & 0.705 & 0.663 & 0.987 & 0.048 & 0.868 & 0.182 & 0.711 \\
            & Ours & {0.180} & \textbf{0.019} & \textbf{0.138} & \textbf{0.344} & \textbf{0.019} & \textbf{0.323} & {0.044} & \textbf{0.062} & \textbf{0.018} & \textbf{0.127} \\
            \midrule
        \end{tabular}
    }
    \label{table:TUMmono}
\end{table*}

\begin{table*}[t]
    \caption{
        \textbf{Stereo SLAM results on EuRoC Dataset with Metric: ATE[m]}. 
    }
    \centering
    \footnotesize
    \resizebox{\linewidth}{!}{%
        \begin{tabular}{cl| ccccc | ccc | ccc | c}
            & Method & MH01 & MH02 & MH03 & MH04 & MH05 & V101 & V102 & V103 & V201 & V202 & V203 & Avg \\
            \toprule
            \multirow{2}{*}{\rotatebox[origin=c]{0}{Sup}} 
            & D3VO + DSO~\cite{d3vo} & - & - & 0.08 \ \ & - & 0.09 \ \ & - & - & 0.11 \ \ & - & 0.05 \ \ & - & - \\
            & DROID-SLAM*~\cite{teed2021droid} & \textbf{0.015} & \textbf{0.013} & \textbf{0.035} & \textbf{0.048} & \textbf{0.040} & \textbf{0.037} & \textbf{0.011} & \textbf{0.020} & \textbf{0.018} & \textbf{0.015} & \textbf{0.017} & \textbf{0.024} \\
            \midrule
            \multirow{3}{*}{\rotatebox[origin=c]{0}{Trad}} 
            & VINS-Fusion~\cite{vinsf} & 0.540 & 0.460 & 0.330 & 0.780 & 0.500 & 0.550 & 0.230 & - & 0.230 & 0.200 & - & - \\
            & SVO~\cite{svo} & {0.040} & 0.070 & 0.270 & 0.170 & 0.120 & {0.040} & {0.040} & {0.070} & {0.050} & 0.090 & 0.790 & 0.159 \\
            & ORB-SLAM3~\cite{orbslam3} & \textbf{0.029} & \textbf{0.019} & \textbf{0.024} & \textbf{0.085} & \textbf{0.052} & \textbf{0.035} & \textbf{0.025} & \textbf{0.061} & \textbf{0.041} & \textbf{0.028} & \textbf{0.521} & \textbf{0.084} \\
            \midrule
            \multirow{2}{*}{\rotatebox[origin=c]{0}{S-Sup}} 
            & Dynamo-Depth (Mono) \cite{sun2024dynamo} & 4.26 & 3.77 & 3.43 & 6.30 & 6.46 & 1.76 & 1.75 & 1.48 & 2.02 & 2.05 & 1.91 & 3.16 \\
            & Ours & \textbf{0.164} & \textbf{0.140} & \textbf{0.202} & \textbf{0.299} & \textbf{0.257} & \textbf{0.076} & \textbf{0.054} & \textbf{0.116} & \textbf{0.044} & \textbf{0.097} & \textbf{0.246} & \textbf{0.154} \\
            \midrule
        \end{tabular}
    }
    \label{table:EurocStereo}
\end{table*}

\begin{table*}[t]
    \centering
    \footnotesize
    \caption{
        \textbf{Stereo SLAM results on TartanAir Stereo Benchmark with Metric: ATE[m]}. 
    }
    \resizebox{1\linewidth}{!}{%
        \begin{tabular}{cl|cccccccc | c}
            & Method & SH000 & SH001 & SH002 & SH003 & SH004 & SH005 & SH006 & SH007 & Avg \\
            \toprule
            \multirow{2}{*}{Sup} & TartanVO~\cite{wang2021tartanvo} & 2.52 & {1.61} & 3.65 & 0.29 & 3.36 & {4.74} & 3.72 & 3.06 & 2.87 \\
            & DROID-SLAM*~\cite{teed2021droid} & \textbf{0.44} & \textbf{0.08} & \textbf{0.13} & \textbf{0.20}  & \textbf{0.16} & \textbf{3.29} & \textbf{0.38} & \textbf{0.18} & \textbf{0.61} \\
            \midrule
            Trad & ORB-SLAM2~\cite{orbslam2} & 0.05 & 6.67 & X & X & X & X & 0.10 & X & - \\
            \midrule
            \multirow{2}{*}{S-Sup} & Dynamo-Depth (Mono) \cite{sun2024dynamo} & 37.73 & 31.11 & 14.38 & 11.32 & 14.39 & 29.43 & 46.85 & 24.51 & 26.71 \\
            & Ours & \textbf{0.32} & \textbf{1.94} & \textbf{1.26} & \textbf{0.15} & \textbf{0.29} & \textbf{6.7}2 & \textbf{0.33} & \textbf{0.08} & \textbf{1.39} \\
            \bottomrule
        \end{tabular}
    }
    \label{table:TartanAir stereo}
\end{table*}

\begin{figure*}[t]
  \centering
  \includegraphics[width=\linewidth]{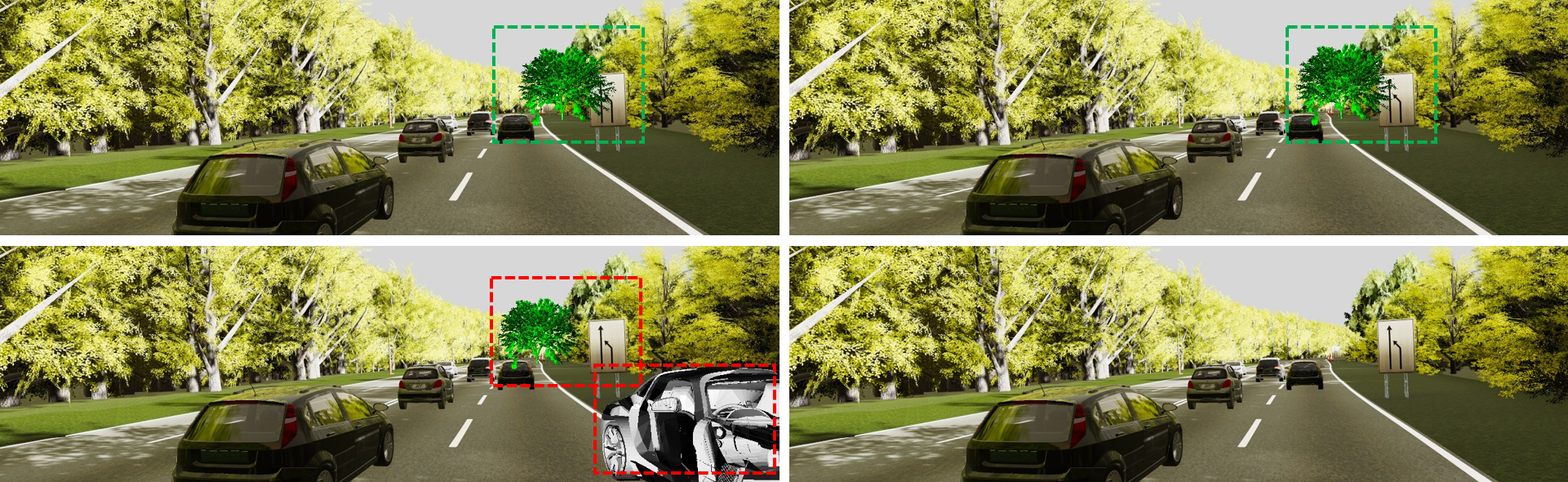}
  \caption{\textbf{AR Application.} We augment the original video with a virtual tree and a car. From left to right and from top to down: Ground-Truth, D$^3$FlowSLAM, DROID-SLAM, Original Image. }
  \label{fig:AR demo}
\end{figure*}

\begin{figure*}[t]
    \centering
    \includegraphics[width=1\linewidth]{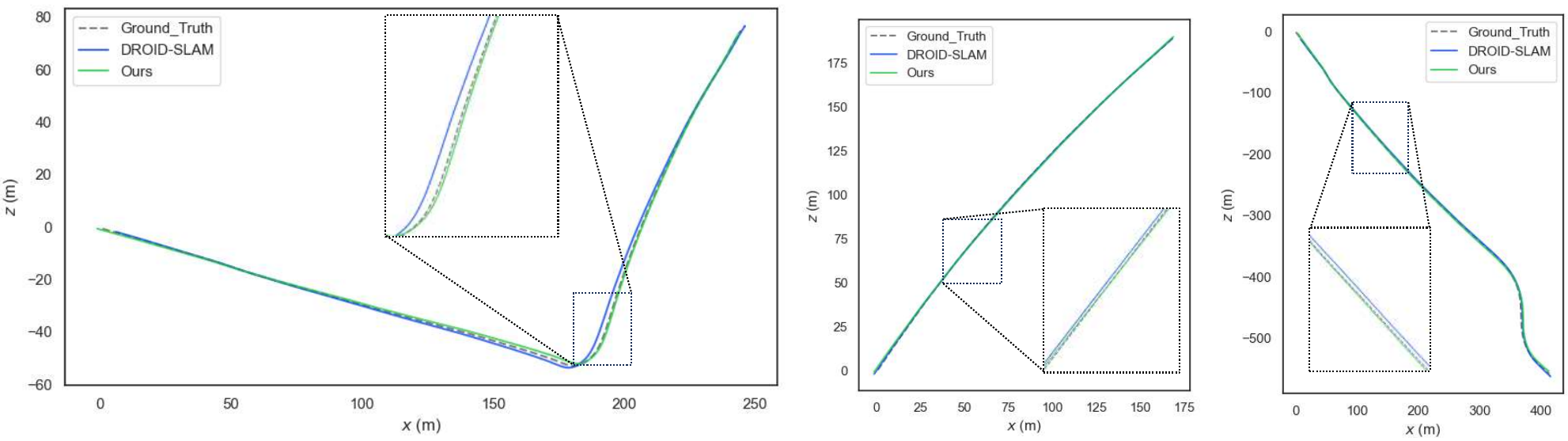}
    \caption{\textbf{Comparison Trajectory Results of Our Method with DROID-SLAM.} In dynamic sequences like VKITTI2 Sequences 01(left), 18(middle), and 20(right), our method performs better than DROID-SLAM, with better trajectory estimation results.}
    \label{fig: trajetory of vkitti2}
\end{figure*}

\begin{figure*}[t]
    \centering
    \includegraphics[width=\linewidth]{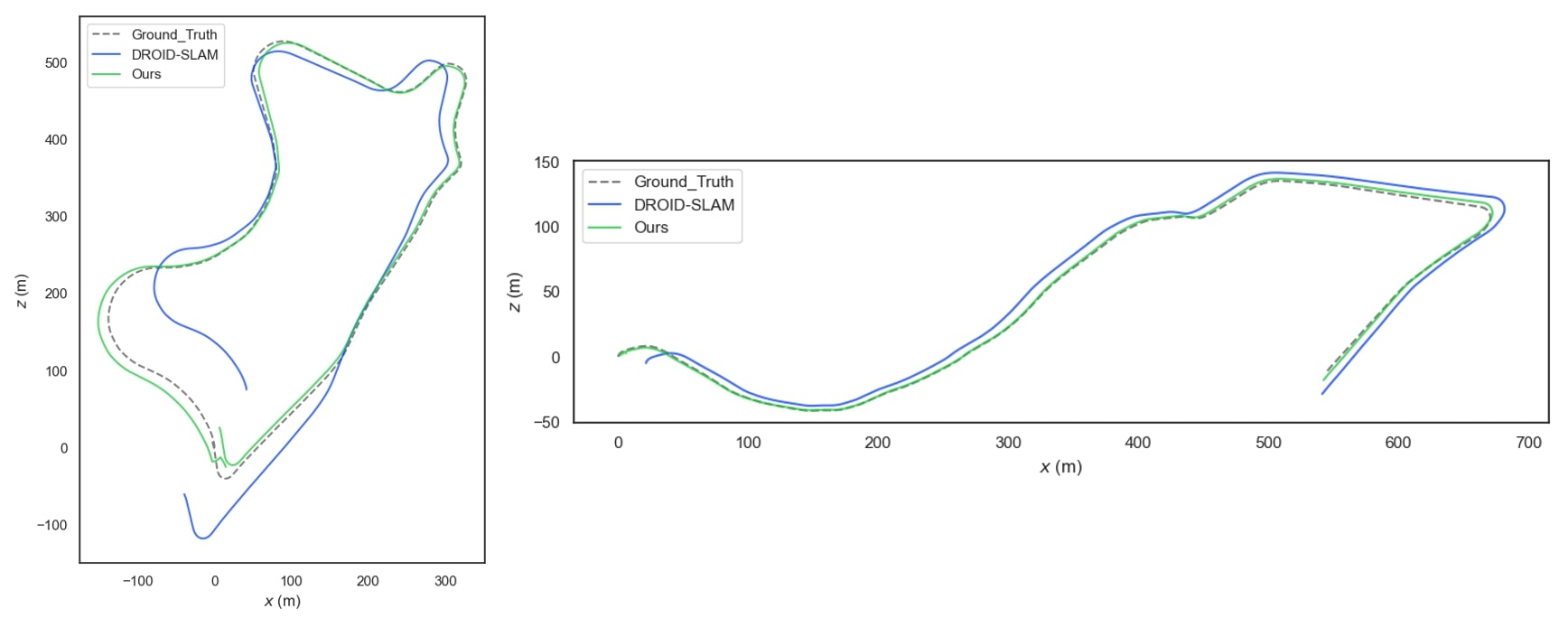}
    \caption{\textbf{Trajectory Comparison between Our Method and DROID-SLAM.} In KITTI sequences 09(Left) and 10(Right), our trajectories are closer to the ground truth.}
    \label{fig: trajetory of kitti}
\end{figure*}

\subsection{Update Module}
\ywc{
DROID-SLAM introduces an update operator where a $3\times3$ ConvGRU is used to update the hidden state $\mathbf{h}$, the camera pose $\mathbf{G}$ 
and depth $\mathbf{d}$. 
The final pose and depth are obtained by applying incremental updates $\Delta \boldsymbol \xi^{(k)}$ and $\Delta \mathbf{d}^{(k)}$ to the current estimates through retraction on the SE(3) manifold and vector addition, respectively.
}:
\begin{equation}
    \mathbf{G}^{(k+1)} = \Exp(\Delta \boldsymbol \xi^{(k)}) \circ \mathbf{G}^{(k)}, \qquad
    \mathbf{d}^{(k+1)} = \Delta \mathbf{d}^{(k)} + \mathbf{d}^{(k)}.
\end{equation}
The update operator iteratively produces a sequence of poses and depths with the goal of converging to a fixed point: $\{\mathbf{G}^{(k)}\} \rightarrow \mathbf{G}^*$, $\{\mathbf{d}^{(k)}\} \rightarrow \mathbf{d}^*$.

\section{DINO Guidance}
For DINO Guidance, we have also considered using features from the DINOv2 \cite{oquab2023dinov2} model. However, our basic requirement for foreground masks is precision, with finer edges for foreground objects being preferable. Despite DINOv2 providing higher-quality features compared to DINO \cite{caron2021emerging}, its lower resolution feature maps lead to poorer mask quality when using clustering methods \cite{amir2021deep}, resulting in less detailed masks. Based on this, we ultimately choose DINO for DINO Guidance.

\section{Implementation Details}
\subsection{Training Initialization}
To improve training efficiency, we preprocess all training data using functions provided by OpenCV \cite{bradski2000opencv} to select image pairs. We assume that the training data is temporally continuous. For any given frame, we consider the $N$ adjacent frames as the selection window. We first filter out frames in this window that can be triangulated with the current frame. Then, we use OpenCV to compute sparse optical flow from the remaining matched points to the current frame. Frames with an average optical flow outside a specified range are discarded. The remaining frames are considered candidates, and during training, we randomly select one of these frames to pair with the current frame.

\subsection{SLAM System Details} 
We follow similar system settings as described in \cite{teed2021droid}.
During initialization, D$^3$FlowSLAM continuously receives new frames until a total of 12 are collected. It constructs frame graphs for these frames and uses our dynamic update module to compute their initial pose and inverse depth maps.
In the front-end, when a new frame arrives, the system constructs a temporary graph with the new frame and its three nearest neighbors. Within this graph, the hidden states of the new frame, such as pose and inverse depth, are optimized.
In the back-end, the system creates a new graph containing all preserved keyframes. Edges between keyframes are generated according to specific rules to eliminate redundancy. The dynamic update module is then used to optimize the entire graph for final poses and depths.


\section{Experiments}


\subsection{More SLAM Results}
The detailed results for TUM-RGBD static sequences are presented in Tab.~\ref{table:TUMmono}. Our method outperforms the latest self-supervised method and matches the performance of supervised and traditional methods, achieving an average ATE of 0.127m.
We also test our system in the stereo input mode.
We test our method on TartanAir-Test stereo dataset and EuRoC stereo dataset. The self-supervised method \cite{sun2024dynamo} we compare does not support stereo input, so we use its monocular results and indicate this in the tables. 
Tab.~\ref{table:EurocStereo} and Tab.~\ref{table:TartanAir stereo} illustrate that our method is better than the latest self-supervised method. It also achieves results comparable to some traditional and supervised methods, with an average ATE of 1.387m on the TartanAir-Test stereo dataset and 0.154m on the EuRoC stereo dataset.

\subsection{Finetuning Settings}
As mentioned in the main paper, we finetune our pretrained model, which was initially trained on the full TartanAir and PointOdyssey datasets, using the Virtual KITTI2, KITTI, TartanAir-Test, and EuRoC datasets. Since our method is self-supervised, we only require images from these sequences for finetuning, with a learning rate of 0.0001. For Virtual KITTI2, we use the \textit{clone} split of all scenes. For KITTI, we use Seq \textit{00, 01, 07, 08}. For TartanAir-Mono-Test, we use \textit{ME003, MH006}. For EuRoC, we use right(left for evaluation) camera images for Seq \textit{MH01, MH02, V101, V201}. The finetuning images for Dynamo-Depth \cite{sun2024dynamo} remain the same.

Tab.~\ref{table:finetune comp} shows the results for part of sequences from the EuRoC and TartanAir-Test datasets before and after finetuning. It shows that the finetuning significantly enhances performance on sequences where the initial results were suboptimal. This underscores our model's ability to adapt to new datasets and environments using a self-supervised scheme.

\begin{table}[t]
    \centering
    \caption{
        \textbf{The Comparison of our method's results Before and After Finetuning with Metric: ATE[m].} We present the evaluation ATE results for part of sequences from the EuRoC and TartanAir-Test datasets. The finetuning operation helps correct some significantly abnormal results.
    }
    \footnotesize
    \resizebox{1.0\linewidth}{!}{%
        \begin{tabular}{l|cccc|cc}
            \multicolumn{1}{r|}{Dataset} & \multicolumn{4}{c|}{EuRoC} & \multicolumn{2}{c}{TartanAir-Test} \\
            \midrule
            Method & MH02 & MH03 & V101 & V201 & MH000 & MH006 \\
            \midrule
            Ours (raw) & 0.281 & 1.789 & 1.183 & 1.028 & 1.380 & 24.507 \\
            Ours (finetuned) & 0.184 & 0.210 & 0.058 & 0.041 & 0.833 & 0.194 \\
            \bottomrule
        \end{tabular}
    }
    \label{table:finetune comp}
\end{table}

\subsection{Trajectory Visualization}


We present some of our trajectory visualization results to intuitively demonstrate the effectiveness of our trajectory estimation. In Fig.~\ref{fig: trajetory of vkitti2} we show the trajectories in the Virtual KITTI2 dataset. In Fig.~\ref{fig: trajetory of kitti} we show the trajectories in the KITTI  dataset. As seen in these figures, our trajectories in these dynamic scenes are closer to the ground truth trajectories compared to other approaches.




\subsection{AR Applications}


We conduct extensive experiments on AR applications in dynamic scenes to demonstrate the robustness of our method. In Fig.~\ref{fig:AR demo}, we augment the original video with virtual elements such as a tree, car, and street lamp. D$^3$FlowSLAM effectively handles dynamic objects in the scene, while other methods, like \cite{teed2021droid}, show drift, especially in areas highlighted by the red box.

\section{Discussion and Limitations}
We propose D$^3$FlowSLAM, a self-supervised approach that significantly outperforms \textbf{existing self-supervised methods} in both dynamic and static scenes. While our method is versatile and robust in dynamic environments, there are areas for improvement. D$^3$FlowSLAM is less effective in certain static scenarios compared to the SOTA supervised methods due to the absence of strong priors like depth and pose during training. To address this, our future work will explore additional supervisory terms or loss functions to enhance self-supervision. We also aim to better integrate powerful self-supervised pretrained models into the entire SLAM system.
Our method currently focuses on camera pose estimation, with depth and optical flow results limited to 1/8 of the original image resolution, which is not ideal for tasks requiring precise depth or optical flow. The model also faces GPU memory limitations with ultra-long image sequences and large scenes, and its running speed does not yet meet real-time requirements. Developing a lightweight, high-performance dynamic SLAM system is a key direction for our future research.


\end{document}